\documentclass[conference,letterpaper,10pt]{ieeeconf}
\IEEEoverridecommandlockouts
\usepackage{epsfig}
\usepackage{amsmath} 
\usepackage{amssymb}  
\usepackage{cite}
\usepackage{epstopdf}
\usepackage{caption}
\usepackage{subcaption}
\usepackage{lscape}
\usepackage{hyperref}
\usepackage[dvipsnames]{xcolor}
\usepackage{color}
\usepackage{url}
\usepackage{array}
\usepackage{slashbox}
\usepackage{cleveref}
\newcolumntype{C}[1]{>{\centering\arraybackslash}p{#1}}

\newtheorem{remark}{Remark}

\newtheorem{exa}{Example}
\newtheorem{prop}{Proposition}
\definecolor{teal}{RGB}{0, 161, 115}
\definecolor{brightBlue}{RGB}{0, 114, 207}
\definecolor{wine}{RGB}{207, 0, 48}
\definecolor{darkgrey}{RGB}{60, 60, 60}

\newcommand{\yA}{y_\mathrm{E}}         
\newcommand{\yB}{y_\mathrm{C}}         
\newcommand{\vA}{v_\mathrm{E}}         
\newcommand{\vB}{v_\mathrm{C}}         
\newcommand{\psiA}{\psi_\mathrm{E}}         
\newcommand{\psiB}{\psi_\mathrm{C}}         
\newcommand{\aA}{a_\mathrm{E}}         
\newcommand{\aB}{a_\mathrm{C}}         
\newcommand{\deltaA}{\delta_\mathrm{E}}         
\newcommand{\deltaB}{\delta_\mathrm{C}}         
\newcommand{\LA}{d_\mathrm{E}}         
\newcommand{\LB}{d_\mathrm{C}}         

\newcommand{\xrel}{x_\mathrm{R}}         
\newcommand{\yrel}{y_\mathrm{R}}         
\newcommand{\jointstate}{z}
\newcommand{\zA}{z_\mathrm{E}}
\newcommand{\ZA}{\mathcal{Z}_\mathrm{E}}
\newcommand{\zB}{z_\mathrm{C}}

\newcommand{\VA}{V_\mathrm{E}}
\newcommand{\fA}{f_\mathrm{E}}
\newcommand{\CA}{\mathcal{C}_\mathrm{E}}
\newcommand{\GA}{\mathcal{G}_\mathrm{E}}

\newcommand{\uA}{u_\mathrm{E}}    
\newcommand{\uB}{u_\mathrm{C}}    

\newcommand{\UA}{\mathcal{U}^\mathrm{E}}    
\newcommand{\UB}{\mathcal{U}^\mathrm{C}}    

\begin{document}

\title{Refining Obstacle Perception Safety Zones \\ via  Maneuver-Based Decomposition}

\author{Sever Topan*, Yuxiao Chen*, Edward Schmerling, Karen Leung,\\ Jonas Nilsson, Michael Cox, Marco Pavone
\thanks{The authors are with NVIDIA, * indicates equal contribution. {\tt\small \{stopan, yuxiaoc, eschmerling, kaleung, jonasn, mbc, mpavone\}@nvidia.com}}}

\newcommand{\klnote}[1]{\textcolor{blue} {KL:#1}}
\newcommand{\stnote}[1]{\textcolor{red} {ST:#1}}

\maketitle

\begin{abstract}
A critical task for developing safe autonomous driving stacks is to determine whether an obstacle is safety-critical, i.e., poses an imminent threat to the autonomous vehicle. Our previous work showed that Hamilton Jacobi reachability theory can be applied to compute \textit{interaction-dynamics-aware} perception safety zones that better inform an ego vehicle's perception module which obstacles are considered safety-critical. For completeness, these zones are typically larger than absolutely necessary, forcing the perception module to pay attention to a larger collection of objects for the sake of conservatism. As an improvement, we propose a maneuver-based decomposition of our safety zones that leverages information about the ego maneuver to reduce the zone volume. In particular, we propose a ``temporal convolution'' operation that produces safety zones for specific ego maneuvers, thus limiting the ego's behavior to reduce the size of the safety zones. We show with numerical experiments that maneuver-based zones are significantly smaller (up to 76\% size reduction) than the baseline while maintaining completeness.
\end{abstract}

\section{Introduction}

The complexity of vehicle autonomy defies the existence of universally applicable performance metrics and objectives, as well as any simple descriptions of the multitudinous constraints that must be satisfied during operation. Despite these challenges, the need to quantify confidence in such systems as a prerequisite for their safe deployment has motivated decompositional approaches whereby various subsystems of the autonomy stack are separately validated (e.g., towards synthesizing a safety argument for the combined system \cite{iso2018road}). Subdividing validation and verification (V\&V) problems into component-wise performance requirements improves the tractability of both certifying and building safe systems. In this work we consider additional data-dependent decomposition of V\&V, further aiding tractability by defining performance requirements on a scenario-specific basis.

One component where evaluation is seemingly straightforward is object detection which, when regarded in isolation, enjoys a plethora of established metrics from the computer vision community capturing performance on a given dataset. The choice of what this validation data should be, however, invokes ``full-stack'' considerations of which obstacles (e.g., other vehicles, pedestrians) in a scene have the potential to be safety-critical, taking into account downstream behavior prediction, planning, and control components. That is, while we might aspire to demand high detection performance for all obstacles in a large radius around an autonomous vehicle (AV), in practice perception system performance should be optimized for a more restricted, task-specific \emph{perception zone}. Prior works have defined such AV perception zones on the basis of predictive assumptions on agent behavior \cite{volk2020comprehensive,bansal2021risk} or otherwise first-principles reachability analysis considering the underlying dynamics of the AV-obstacle interaction \cite{Topan2022Interaction}.

\begin{figure}
    \centering
    \includegraphics[width=0.45\textwidth]{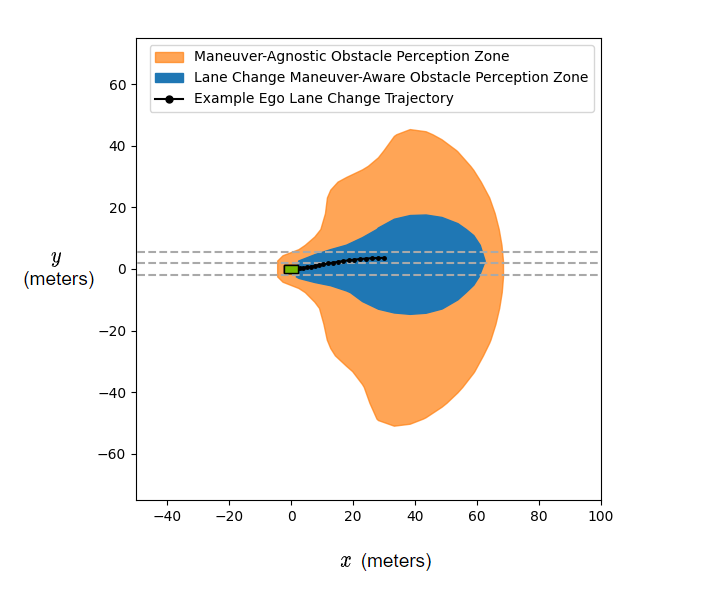}
    \caption{Incorporating \textit{maneuver-based decomposition} in obstacle perception zone calculation can significantly improve soundness while maintaining desirable completeness properties. We demonstrate how constraining ego vehicle dynamics to a family of lane change maneuvers during zone computation (blue) can minimize volume when compared to an unconstrained example (orange). An example of a valid lane change considered within the maneuver constraint is shown in black.}
    \label{fig:hero}
\end{figure}

In order to weigh these approaches against each other, it is useful to introduce notions of \emph{completeness} and \emph{soundness}, both desirable properties of perception zones. Informally, a zone is complete if all safety-critical objects in a scene are captured within; a zone is sound if all objects within the extent of the zone indeed have the potential to be safety-critical. Of these two, completeness is non-negotiable: it is necessary that a perception system raise all obstacles relevant to AV safety for downstream consideration/computation. Soundness requirements are therefore typically relaxed, e.g., as in \cite{Topan2022Interaction}, with the interpretation of adding a conservative expectation for detecting further objects beyond the zones' necessary completeness. However, recognizing that the intent of defining perception zones is to make the development of valid object detectors more tractable, we seek to improve the soundness of our zone constructions.

In this work we examine how conditioning on the specific driving scenario, i.e., computing perception metrics specific to an operational design domain (ODD), may be applied to improve soundness without sacrificing completeness. This may be regarded as a natural extension of \cite{Topan2022Interaction}. In that work knowledge of interaction dynamics, representing downstream behavior planning subsystems, was used to restrict the designation of obstacles as safety-critical within interaction-dynamics-aware perception zones, or \emph{safety zones} for short. Here we leverage the additional context of the AV's intended driving maneuver to further reduce safety zone size and improve V\&V tractability. Computationally we continue to apply tools from Hamilton-Jacobi (HJ) reachability analysis \cite{mitchell2005time,bansal2017hamilton} to completely account for the closed-loop interaction dynamics between the AV and obstacle agents, now with additional constraints specifying AV maneuver completion. Accounting for such constraints is the primary technical challenge addressed by this work. That is, compared to previous safety zones \cite{Topan2022Interaction} that consider only the AV/obstacle interaction up until potential collision within a time horizon, these refined \emph{maneuver-aware} perception zones require this potential collision to occur intermediately along one of the ways the AV planner may choose to accomplish the maneuver.

\noindent \textbf{Statement of Contributions:} Our contributions in this work are two-fold, specifically:
\begin{enumerate}
    \item We refine the interaction-dynamics-aware perception zones put forward in \cite{Topan2022Interaction} by equipping HJ-reachability formulations with AV maneuver constraints, thus producing tighter maneuver-conditioned zones without introducing safety blind spots. The computation of these maneuver-aware perception zones is enabled by a novel ``temporal convolution'' operation which enables modeling of temporal dependencies between state goals, i.e., potential collision conditioned on the ego maneuver completion.
    \item We instantiate the maneuver-aware perception zones for two representative scenarios: a lane change and a right-angle turn, and conduct a series of experiments that demonstrate that these constraints significantly improve zone soundness while maintaining completeness.
\end{enumerate}
We believe that the resulting perception safety zones may be used in a ``drop-in'' fashion, sub-selecting objects to consider for AV perception metrics computation, in order to direct development effort towards the most impactful increases in AV performance, safety, and trust.

\section{Related Work}

Computing a dynamics-aware perception zone can be formulated as a reachability problem \cite{Topan2022Interaction}, and in this section, we discuss reachability-based approaches that trade off between completeness and soundness.
Reachability analysis \cite{majumdar2017funnel,bansal2017hamilton,girard2005reachability,kurzhanski2000ellipsoidal,yazarel2004geometric,althoff2008reachability} is a popular verification technique for ensuring a, possibly stochastic, system stays within a desired range of operation and does not reach forbidden regions in the state space.
In safety-critical human-robot interactive settings, such as in autonomous driving, to make strong safety arguments, oftentimes overly conservative assumptions are made to account for uncertainty in human behaviors. However, such assumptions may result in impractically sized sets that unduly impede robot mobility (e.g., frozen robot problem \cite{trautman2010unfreezing}).

HJ reachability is a mathematical formalism for characterizing the performance and safety properties of (multi-agent) dynamical systems \cite{mitchell2005time,margellos2011hamilton,bansal2017hamilton}.
Typically a single reachable set is computed offline and is formulated to account for all possible, including worst/best-case, agent's (human or robot) policies regardless of the agent's high-level intent \cite{leung2020infusing,wang2020infusing}. Since all possible agent policies are considered, the resulting reachable set is complete \cite{Topan2022Interaction}. However, through this HJ-based formulation, the reachable set becomes a ``one-set-fits-all-scenarios'' set, possibly leading to an impractically large set.

That is, when using the set in any particular ODD, the reachable set will be accounting for behaviors that may be unrealistic, or out-of-scope, for that ODD.

To tackle the issue of over-conservatism and compute reachable sets that are more practically-sized, other approaches restrict the behaviors of agents, such as assuming agents will stay in their lane \cite{althoff2014online,althoff2011setbased}, assuming all agents will brake in an emergency scenario \cite{shalev2017formal,nister2019safety}, or that all agents will maintain constant velocity \cite{van2008reciprocal}. While such approaches can significantly reduce over-conservatism and hence improve practicality of the set, they are not complete as they neglect to consider other possible behaviors. In the context of computing perception zones for obstacle detection safety evaluation, completeness is a necessary property. In our present work we make assumptions on the intent of AV behavior at the routing level, but allow for any variations in lower-level planning that accomplish this behavior. In this context possibly the closest work, algorithmically speaking, to ours is \cite{chen2018signal} which allows for modeling temporal dependencies (e.g., considering collisions occurring along the way to an AV completing its maneuver) in an HJ formulation. Compared to \cite{chen2018signal}, the temporal convolution operation we advance in this work accounts for the necessary temporal dependency without requiring any additional state dimensions; keeping the dimensionality as low as possible is paramount for the computational tractability of HJ-based approaches.

Alternatively, recent works have turned to \textit{online estimation} of human intent to adapt the reachable set to correspond to the estimated human intent model \cite{tian2022safety,nakamura2022online,borquez2022online}, and therefore improve the soundness of the reachable set. However, these methods primarily focus on estimating parameters describing a human behavior prediction model to help reduce the over-conservatism in an AV's planning algorithm. However, utilizing human behavior prediction is incompatible with our goal of computing perception zones for obstacle detection safety evaluation since we require the computation of perception zones to be modular and independent of other modules in the autonomy stack.

\section{Preliminaries and Notation}
In this section, we briefly introduce the Hamilton-Jacobi-Bellman (HJB) equation which is the mathematical formulation underpinning the computation of our perception zones, and establish some notation used throughout this paper.
\subsection{Hamilton-Jacobi-Bellman Equation}
In this section, we provide a short introduction to the optimal control theory and the HJB equation which is central to the computation of perception safety zones studied in this work. Since a perception zone describes regions where it is possible for two vehicles to collide (under assumptions about their behaviors), we can check if collision is possible by solving an optimal control problem.

An optimal control problem deals with the problem of finding a control law $u(\cdot)$ for a given system that minimizes a cost functional subject to state, control, and dynamics constraints.
Let the dynamics of a system be $\dot{x} = f(x,u,t)$ where $x\in\mathbb{R}^n$ is the state, and $u\in\mathcal{U}\subset\mathbb{R}^m$ is the control input. For a single agent case, $x$ and $u$ denotes the agent's state and controls respectively, and $f$ the agent's dynamics. 
In the case of two agents interacting with each other, $x$ would denote the joint (or relative) state between the two agents, $u$ would describe the joint controls, and $f$ would denote the joint (or relative) dynamics. 

Consider a cost function $J(x(\cdot), u(\cdot), t_0, t_\mathrm{f}) = D(x(t_\mathrm{f})) + \int_{t_0}^{t_\mathrm{f}} C(x(t),u(t),t)dt$ where $D:\mathbb{R}^n \rightarrow \mathbb{R}$ is the terminal cost (e.g., distance to final goal) and $C:\mathbb{R}^n \times \mathbb{R}^m \times \mathbb{R} \rightarrow \mathbb{R}$ is the running cost (e.g., control effort) over the time horizon $[t_0, t_\mathrm{f}]$. For the current time $t_0$, let the current state of the system be $x(t_0) = x_0$. Then the optimal control law $u(\cdot)$ that minimizes the cost function $J$ over a time horizon $[t_0, t_\mathrm{f}]$ is the solution to the following optimization problem,

\begin{eqnarray}
\min_{u} &&  D(x(t_\mathrm{f})) + \int_{t_0}^{t_\mathrm{f}} C(x(t),u(t),t)dt \notag\\
\text{s.t.}&& \dot{x} = f(x,u,t)\label{eq:optimal control problem}\\
& &x(t_0) = x_0\notag\\
&& u(t) \in \mathcal{U}\notag
\end{eqnarray}

To solve for the optimal control law $u^\star(\cdot)$ from \eqref{eq:optimal control problem} for all initial states, it can be shown that it is equivalent to solving the HJB partial differential equation (PDE),

\begin{equation}
    \frac{\partial V(x,t)}{\partial t} + \min_{u\in\mathcal{U}} \left\{\nabla_x V(x,t)^\intercal f(x,u,t) + C(x,u,t) \right\}=0\\
    \label{eq:HJB}
\end{equation}
where $V(x,t_\mathrm{f}) = D(x)$. 
The solution to the HJB PDE $V(x,t)$ is known as the Bellman value function which represents the cost incurred from starting at state $x$ at time $t$ and controlling the system optimally (i.e., with $u^\star(x) = \min_u \left\{ \frac{\partial V(x,t)}{\partial x}^Tf(x,u,t) + C(x,u,t) \right\}$) until time $t_\mathrm{f}$. That is, $V(x,t) = J(x(\cdot), u^\star(\cdot), t, t_\mathrm{f})$. There are three things to note:

\noindent {\bf Zero running cost.} If we assume zero running cost, i.e., $C(x,u,t) = 0 \; \forall x, u, t$, and let $D$ describe the signed distance from a goal region $\mathcal{C}$ (i.e., $\mathcal{C}=\lbrace x \mid D(x) \leq 0 \rbrace$), then the optimal control problem can be interpreted as a reachability problem. 
The value function $V(x,t) = D(x(t_\mathrm{f}))$ is then precisely the terminal cost of the final state if the system currently at state $x$ were to follow the optimal control law $u^\star(\cdot)$ until $t_\mathrm{f}$.
Then $V(x,t) < 0$ indicates that is it possible for the system to be inside the target set at time $t_\mathrm{f}$. 
Moving forward, we will assume zero running cost for the rest of the paper.

\noindent {\bf Two agent system.} In the case of two agents, agent $\mathrm{E}$ and $\mathrm{C}$, let $x$ denote their joint state, $u=[u_\mathrm{E}, u_\mathrm{C}]\in\mathcal{U}^\mathrm{E}\times \mathcal{U}^\mathrm{C}$ their joint control, and $f$ the joint dynamics. Then we can rewrite \eqref{eq:HJB} (with zero running cost) to account for the fact that there are two control input variables to optimize over,

\begin{equation}
    \frac{\partial V(x,t)}{\partial t} + \min_{u_\mathrm{E}\in\mathcal{U}^\mathrm{E}}\min_{u_\mathrm{C}\in\mathcal{U}^\mathrm{C}} \left\{ \nabla_x V(x,t)^\intercal f(x,u_\mathrm{E},u_\mathrm{C},t)  \right\}=0.\\
    \label{eq:HJB two agents}
\end{equation}

\noindent {\bf Reaching target set at any time.} The formulation thus far is concerned with reaching the target set $\mathcal{C}$ at exactly $t_\mathrm{f}$. However, sometimes it is also important consider whether the system enters the target set any time between $[t_0, t_\mathrm{f}]$. To compute the corresponding value function describing entry into the start set at \textit{any} $t\in[t_0, t_\mathrm{f}]$, we can solve a slight variation of \eqref{eq:HJB} (with zero running cost),

\begin{equation}
    \frac{\partial V(x,t)}{\partial t} + \min \left\{ 0, \min_{u\in\mathcal{U}} \left\{\nabla_x V(x,t)^\intercal f(x,u,t)\right\}\right\}=0.\\
    \label{eq:HJB tube}
\end{equation}
We can analogously formulate the corresponding two agent setup for this case too.

We have just described several variations of the HJB equation and each of these variations will be used in certain ways to construct our maneuver-based perception safety zones described in Section~\ref{sec:maneuver-based perception zones}.

\subsection{Notation}
In the rest of the paper, we will refer to autonomous vehicle as the ``ego'' vehicle, and variables corresponding to the ego vehicle will have a subscript $\mathrm{E}$. Similarly, for the (uncontrolled) vehicle that the ego vehicle has detected and wants to check if it is in its perception safety zone, we refer to this uncontrolled vehicle as ``the contender''. Variables corresponding to the contender vehicle will be denoted by a subscript $\mathrm{C}$.
States without any subscripts are assumed to denote the joint state describing both the ego and contender agents. 

\section{Maneuver-based Perception Safety Zones}
\label{sec:maneuver-based perception zones}
In this section, we introduce and motivate the idea of decomposing perception safety zones based on the maneuver type, and then describe the mathematical formulation to construct such a maneuver-based perception safety zone. 

\subsection{High-Level Intuition}

As shown in our previous work \cite{Topan2022Interaction}, safety zones can be calculated with HJ reachability using a ``min-min'' formulation where both the ego vehicle and the contender seek collision. This setup is used to account for the worst-case ego-contender interaction dynamics so the zone can be argued as complete. Due to the conservative modeling constraints used in the prior work, zone geometries can become exessively large as we consider longer time horizons. While it may be warranted to assume that the contender can perform any dynamically feasible maneuvers for the sake of conservatism, the ego vehicle's next high-level maneuver is typically known within its autonomy stack (e.g., lane change, or a right turn). The key idea behind our work is to leverage this information and constrain the obstacle perception zone derivation in order to improve its soundness while retaining completeness.

To help illustrate the idea behind a maneuver-based perception safety zone, consider the following two examples.
\begin{exa}\label{ex:lc}
The ego vehicle's next maneuver is a lane change, which is characterized as the following constraint $\CA=\{\yA\in[y_{\mathrm{des}}-\overline{\delta y},y_{\mathrm{des}}+\overline{\delta y}]\}$. It specifies that the lateral position of the ego vehicle is within $\overline{\delta y}$ meters from the lane center of the desired lane situated at $y_{\mathrm{des}}$. 
\end{exa}

\begin{exa}\label{ex:turn}
The ego vehicle's next maneuver is $90^\circ$ turn to the left, which is characterized as the following constraint $\CA=\{\psiA\ge \pi/2\}$. 
\end{exa}
We are interested in finding all initial conditions such that the ego vehicle can collide with the contender during the horizon and satisfy the constraint $\CA$ by the end of the horizon.

Formally, given a fixed horizon $T$, an ego maneuver is specified by a constraint set $\CA\subseteq \ZA$. The ego constraint can be enforced either at the end of the horizon or throughout the whole horizon. In Example \ref{ex:lc}, the latter is used, and maneuvers such as turning or overtaking can be defined similarly. The challenge, however, lies in capturing both collision-seeking, and maneuver-completing behavior in all possible permutations. This is challenging because it requires temporal reasoning that is not easily represented within typical HJ reachability formulations. We discuss this further in the next subsection.

\subsection{Computing Perception Safety Zones with HJ Reachability}
We begin by using $\zA$ and $\uA$ to represent the dynamic state and control input of the ego vehicle, $\zB$ and $\uB$ to represent the dynamic state and control input of the contender, and $\jointstate$ to represent their joint state. The actual state and input signals (e.g. $X,Y$ coordinates, acceleration, steering, etc.) depend on how the specific problem is formulated. In particular, when symmetry can be utilized to reduce the state dimension, the HJ computation is simplified.

\begin{figure}
    \centering
    \includegraphics[width=0.45\textwidth]{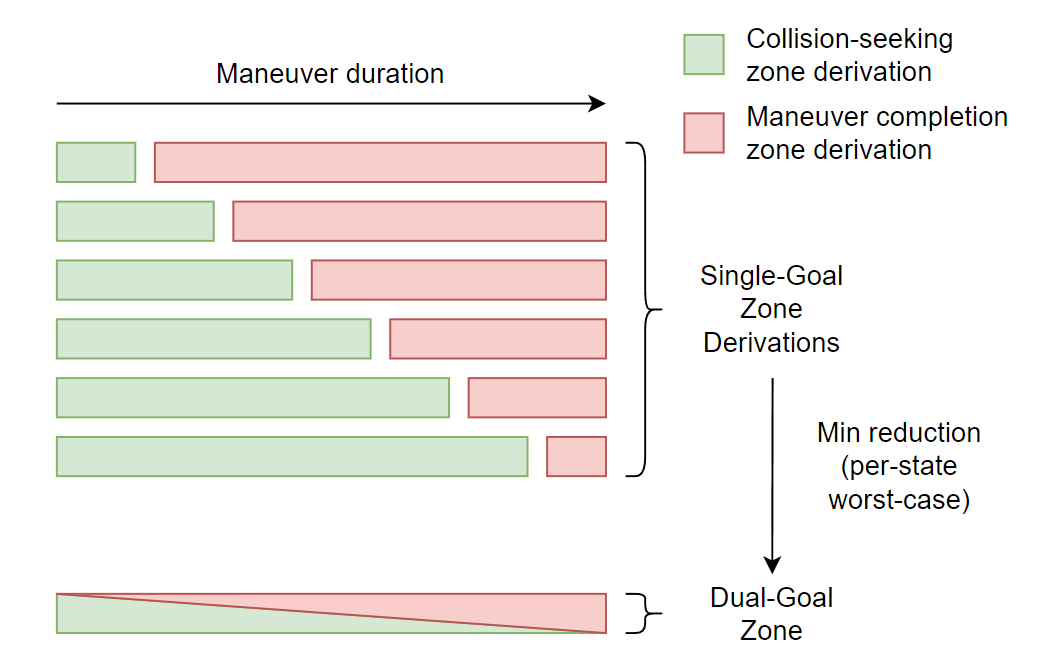}
    \caption{We propose a ``temporal convolution'' approach that can be used to calculate zones with two temporally-related goals. Here, we show how our approach can be used to represent all possible permutations of a temporally-related dual-goal reachability derivation.}
    \label{fig:hero}
\end{figure}
A generic perception safety zone needs to include all initial conditions from which the ego can collide with the contender within the horizon. When ego vehicle's next maneuver is given, we can focus our attention on all possible ways that the ego can finish the maneuver, and ignore any ego behavior that does not satisfy the maneuver specification. The maneuver-based safety zone thus needs to include any initial condition from which the ego vehicle can finish the ego maneuver and incur a collision with the contender in the meantime. 

Such a set in its raw form cannot be directly computed with HJ as it involves two separate tasks (goal reaching and collision seeking) that are not aligned, i.e., the optimal strategy for the two goals are different. To compute the zone, we take a ``convolution''-style approach. In particular, we separate the task into two phases: colliding with the contender at some point within the horizon, and using the remaining time to finish the maneuver.
\begin{remark}
The second phase serves as a certification that the ego is indeed performing the maneuver. If the ego cannot finish the maneuver from the state and time the collision happens, it indicates that the ego behavior in the first phase is not part of a possible realization of the ego maneuver.
\end{remark}

The second phase can be easily computed with a low-dimensional reachability formulation, since only the ego vehicle is involved. For any given time $t<T$, suppose the collision happens at $t$, the second stage can be formulated as a simple reachability computation with boundary condition $\CA$ and horizon $T-t$, where $\CA $ is the goal set. It is well-established that the reachable set can be approximated with the viscosity solution of a HJB PDE \cite{mitchell2005time}:

\begin{align}\label{eq:ego_goal}
    &\VA(\zA,T) = \GA(\zA)\\
    &\frac{\partial \VA(\zA,t)}{\partial t} + \min\{0,\min_{\uA\in\UA} \nabla_{\zA} \VA(\zA,t)^\intercal \fA (\zA,\uA)\}=0 \nonumber
\end{align}
where $\GA(\zA)$ is the boundary value function such that $\CA=\{\zA\mid\GA(\zA)\le 0\}$, $\fA$ is the ego dynamics equation. Once $\VA$ is computed by solving the HJB PDE, $\VA(\zA,t)<0$ indicates that there exists a control strategy that brings $\zA$ to $\CA$ by time $T$, otherwise no such control strategy exists.

Given $\VA$, we notice that once the timing of the collision is fixed and given, we can calculate the reachable set of the following incident: \textit{``the ego vehicle and the contender collide exactly at $t$ and the ego finishes the maneuver before $T$.} The reachable set of this specification can be computed with the following HJB PDE with horizon $[0,t]$:
\begin{align}
    & \mathcal{G}(\jointstate) = \max\{\VA^\uparrow (\jointstate,t), \mathcal{G}_{\mathrm{col}}(\jointstate)\} \label{eq:Vt}\\
    & \frac{\partial V(\jointstate,\tau)}{\partial \tau} + \min_{\uB\in\UB}\min_{\uA\in\UA} \nabla_\jointstate V(\jointstate,\tau)^\intercal f (\jointstate,\uA,\uB)=0 \nonumber
\end{align}
where $\VA^\uparrow$ is the lifting of $\VA$ from $\ZA$ (ego vehicle state space) to $\mathcal{Z}$ (joint state space of ego and contender), that is, any $z\in\mathcal{Z}$ can be projected to $\zA\in\ZA$, and $\VA^\uparrow(z)=\VA(\zA)$. $\mathcal{G}_{\mathrm{col}}$ is a function such that $\mathcal{G}_{\mathrm{col}}(\jointstate)\le0$ if two vehicles are in collision and positive otherwise, $\uB\in\UB$ is the control input of the contender. The above HJ formulation takes the maximum over $\mathcal{G}_{\mathrm{col}}$ and $\VA$ as the boundary condition, and the resulting $V$ satisfies that for $\tau<t<T$, and $\jointstate(\tau)= \jointstate$, if $V(\jointstate,\tau)<0$, there exists a joint strategy between the ego and contender such that the ego and contender collide at time $t$ and the ego is able to finish the maneuver before $T$.

Since $t$ can take any value between 0 and $T$, we sweep through $[0,T]$ and perform multiple HJB PDE computation, and index the resulting value function as $V^t$. $V^t(\jointstate)<0$ indicates that there exists a strategy such that the ego and contender collide at exactly $t$, and the ego is able to finish the maneuver before $T$. Based on simple propositional logic, we have the following proposition.
\begin{prop}
The following two specifications are equivalent: (1) The ego collides with the contender and then finishes the maneuver within $T$. (2) There exists a $t\in[0,T]$ such that the ego and contender collide at $t$, and the ego finishes the maneuver within $[t,T]$.
\end{prop}
We can then compute the maneuver-based zone in the form of a zero-sublevel set of the following value function:
\begin{equation}\label{eq:V_conv}
    V(\jointstate)=\min_{t\in[0,T]} V^t(\jointstate).
\end{equation}
\begin{prop}\label{prop:conv}
Let $V^t$ be the solution of \eqref{eq:Vt}, and $V$ be computed as in \eqref{eq:V_conv}, then for any $\jointstate, V(\jointstate)<0$, there exists a time instance $0\le t\le T$, a control signal of $\uA: [0,T]\to\UA$ and a control signal of $\uB:[0,t]\to\UB$ such that the ego and the contender collide at $t$ and the ego finishes the maneuver before $T$.
\end{prop}
Proposition \ref{prop:conv} follows from HJ reachability theory and the convolution process that we derived above, and is the foundation of our maneuver-based safety zones.

\begin{figure*}[t]
    \centering
    \includegraphics[width=0.65\textwidth]{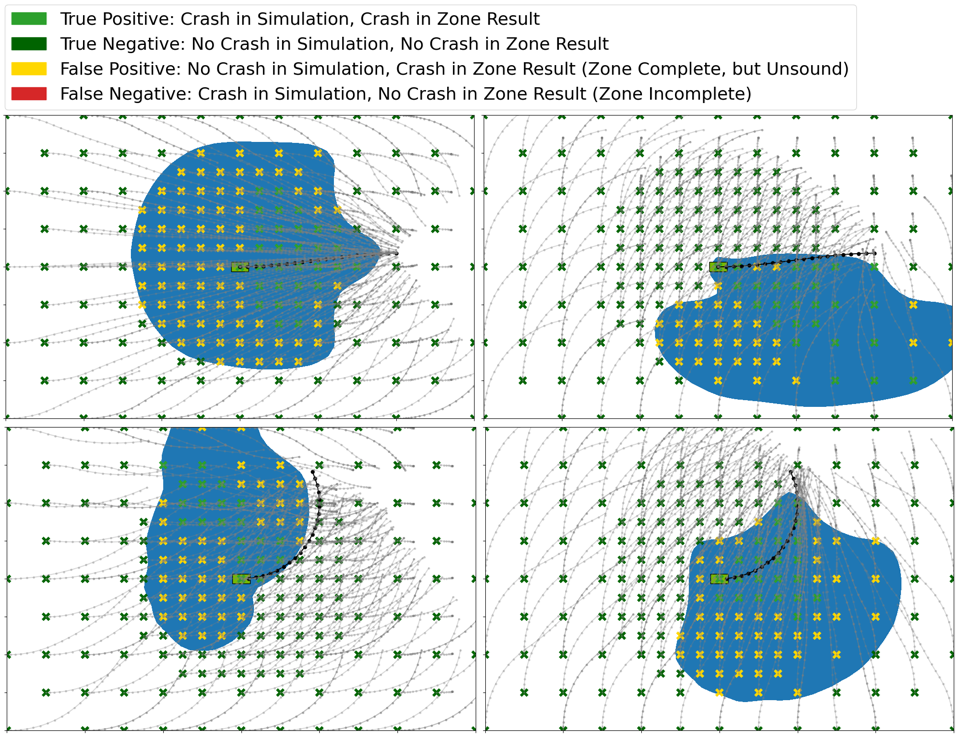}
    \caption{A sampling of experimental results from maneuver-aware obstacle perception zone completeness verification. The obstacle perception zone can be seen in blue, and the simulated trajectory of the ego can be seen in black. Colored markers and gray trajectory lines show trajectories of different contender initial conditions as they simulate a collision attempt with the ego. A green marker indicates that simulation results matched our zone's behavior, while yellow markers show cases where our zone was more conservative than the simulation. We note that this conservatism is expected, since each simulation trial only tests one possible ego trajectory, whereas our zones account for all dynamically feasible trajectories.  \textbf{Top left}: Lane chane maneuver with $\vA$ = 10m/s, $\vB$ = 5m/s, $\psiB$ = 0rad. \textbf{Top right}: Lane change maneuver with  $\vA$ = 10m/s, $\vB$ = 5m/s, $\psiB$ = $\pi/2$rad. \textbf{Bottom left}: $90^{\circ}$ turn maneuver with $\vA$ = 10m/s, $\vB$ = 5m/s, $\psiA$ = 0rad. \textbf{Bottom right}: $90^{\circ}$ turn maenuver with $\vA$ = 10m/s, $\vB$ = 5m/s, $\psiA$ = $\pi/2$rad.}
        \label{fig:experiment-results}
\end{figure*}

\section{Experimental Results}

Our two primary claims are that the introduction of maneuver-awareness i) maintains completeness of our obstacle perception zones while ii) improving soundness. To this end, we first leverage a Model Predictive Controller (MPC) simulation to validate that no combination of feasible initial conditions and ego trajectories result in collisions that were not  identified by our maneuver-aware obstacle perception zone. We then compare state-space volumes of our maneuver based zones against zones derived without maneuver constraints to measure soundness improvements. We begin with a precise definition of our maneuver-based zones, which is described in the next section.

\subsection{Maneuver-Aware Zone Definition} 

Throughout our experiments, we use two prototypical maneuvers: A lane-change, and a rail-based 90-degree turn.

In the lane-change case, we use a 3D model for the ego behavior:
\begin{equation}
    \begin{bmatrix}\dot{\yA}&\dot{\psiA}&\dot{\vA}\end{bmatrix}^\intercal = \begin{bmatrix}
        \vA \sin(\psiA) & \frac{\vA\tan(\deltaA)}{\LA} & \aA
    \end{bmatrix}^\intercal,
\end{equation}
where $\yA,\psiA$, and $\vA$ are the lateral position, heading angle, and velocity of the ego, the inputs are the acceleration $\aA$ and steering $\deltaA$, $\LA$ is the wheel-base of the ego vehicle. The joint dynamics between the ego and the contender is described by a 7D dynamics:
\begin{equation}
    \begin{bmatrix}
        \dot{\xrel}\\
        \dot{\yA} \\
        \dot{\yB} \\
        \dot{\psiA}\\
        \dot{\psiB}\\
        \dot{\vA}\\
        \dot{\vB}
    \end{bmatrix} = \begin{bmatrix}
        \vB\cos(\psiB)-\vA\cos(\psiA)\\
        \vA\sin(\psiA)\\
        \vB\sin(\psiB)\\
        \frac{\vA\tan(\deltaA)}{\LA}\\
        \frac{\vB\tan(\deltaB)}{\LB}\\
        \aA\\
        \aB
    \end{bmatrix},
\end{equation}
where $\xrel$ is the longitudinal distance between the ego and the contender, the contender state and inputs are named in the same way as the go with subscript $\mathrm{C}$.

The Lane-Change maneuver completion condition is defined as in example \ref{ex:lc}, with lane width 3.6 meters and lateral terminal offset of $\overline{\delta Y} = 0.8$. 

The Rail-based maneuver is defined by constraining ego position and heading to a curve of varying curvature (function of heading angle). Maximum ego velocity must be chosen such that the minimum curvature is a feasible at the given maximum speed. For simplicity, we use a fixed turning radius of $R = 20$ meters and maximum Ego speed at 10 meters per second. The ego dynamics is described by a 2D system:
\begin{equation}
    \begin{bmatrix}
        \dot{\psiA}&
        \dot{\vA}
    \end{bmatrix}^\intercal=\begin{bmatrix}
        \vA \kappa(\psiA)&
        \aA
    \end{bmatrix}^\intercal,
\end{equation}
where $\kappa(\cdot)$ is the curvature as a function of the heading angle, which is specified by the rail.
The joint state space between the ego and the contender is described by a 6D dynamic system:
\begin{equation}
    \begin{bmatrix}
        \dot{\xrel}\\
        \dot{\yrel}\\
        \dot{\psiA}\\
        \dot{\psiB}\\
        \dot{\vA}\\
        \dot{\vB}
    \end{bmatrix} = \begin{bmatrix}
        \vB\cos(\psiB)-\vA\cos(\psiA)\\
        \vB\sin(\psiB)-\vA\sin(\psiA)\\
        \vA \kappa(\psiA)\\
        \frac{\vB\tan(\deltaB)}{\LB}\\
        \aA\\
        \aB
    \end{bmatrix}.
\end{equation}
The completion condition is defined by $\mathbf{sign}(\psi_{\text{des}})(\psiA - \psi_{\text{des}})>0$, where $\psi_{\text{des}}$ is target heading.

All zones were computed using a Nvidia RTX 3090. The Lane Change zone took 979 seconds to compute, while the 90-degree turn zone took 465 seconds. Despite an expensive offline derivation, online zone queries are rapid as they entail a simple grid look-up along with an optional interpolation operation.

\subsection{Completeness Verification} \label{sec:completeness-verification}

We begin by verifying our claim that leveraging maneuver-based obstacle perception zones does not compromise completeness of our analysis. Since obstacle perception zones reflect the possibility of a collision given a particular dynamic state, it is not currently feasible to verify their completeness on real data as this would require recording scenarios where real vehicle collisions occur. In the absence of such data, we resort to simulation to demonstrate the completeness of our maneuver-based zones.

We leverage simulations where the ego-vehicle follows a fixed trajectory that is carefully chosen to abide by appropriate dynamic constraints and maneuvers used in obstacle perception zone derivation. Our contender is simulated using an adversarial MPC controller that assumes constant ego velocity at each time step, and pursues a collision with a preview strategy. To select initial conditions, we fix $\psiA$ and $\yA$ where applicable to 0, representing the start of a given maneuver. We then sweep over the remaining dimensions to obtain 22,944 trials in total. The sweep is performed uniformly, with the exception of the $\xrel$, $\yrel$ and $\yB$ coordinates which are sampled more densely closer to the ego vehicle.

We use the following terminology to refer to our results. \textit{True positives} and \textit{true negatives} indicate simulations where both our zone and the simulation observe the same results; Namely, either both or neither observe a collision (respectively). A \textit{false positive} is a scenario where our zone predicts a collision, but none are observed in simulation. Note that an individual trial of our simulation only considers one possible ego-vehicle trajectory, while our reachability-based zones consider the family of all dynamically-feasible ego trajectories together. Moreover, our adversarial MPC controller may not be optimal (worst-case), constituting another source of false-positive cases. False positive results are thus expected from our experiments, and do not compromise our completeness claims. \textit{False negatives}, however, indicate scenarios where our zone failed to predict a collision that occurred in simulation. These results are problematic, as they render our zones incomplete. We do note however, that since we use the same dynamical models between our reachability-based zones and our simulation, some numerical error can be expected near zone boundaries that can result in false negatives. In practice, the dynamical models used in reachability-based zones should be chosen to be conservative approximations.

Our quantitative results can be found in table \ref{table:completeness-experiment}. While we observe no false negatives in our lane change maneuver zone, we do find a small number of failures in our turning maneuver zone. All of the observed failures occur very close to our zone boundary, occurring at an average value of 0.34 and a maximum of 1.34 (recall that the safety zone is computed as the zero-sublevel set of the value function $V$). These values are small in the sense that all false negatives could be made to lie within the zone, becoming true positives, by simply inflating the threshold for ego/contender collision by a corresponding 1.34 meters (approximately half a car width). We report these false negatives to illustrate the fact that in practice, though our theory guarantees completeness, error stemming from, e.g., coarse grid discretization in the numerical PDE solver, requires some additional safety buffer.

\definecolor{tpcolor}{HTML}{2ca02c}
\definecolor{tncolor}{HTML}{006400}
\definecolor{fpcolor}{HTML}{ffd700}
\definecolor{fncolor}{HTML}{d62728}

\begin{table}[h]
\centering
\begin{tabular}{ |c | c | c | c | c | c | }
    \hline 
    Zone & \textcolor{tpcolor}{$\mathbf{\times}$} True & \textcolor{tncolor}{$\mathbf{\times}$} True & \textcolor{fpcolor}{$\mathbf{\times}$} False & \textcolor{fncolor}{$\mathbf{\times}$} False \\ 
    Name & Positive & Negative & Positive & Negative \\ 
    \hline 
    \hline
    Lane Change & 12.4\% & 59.5\% & 18.1\% & 0.0\% \\  
    $90^{\circ}$ Turn & 8.8\% & 75.1\% & 15.9\% & 0.2\% \\  
    \hline 
\end{tabular}
\caption{Experimental results from our completeness verification. See section \ref{sec:completeness-verification} for an explanation of terminology used in column names. We observe high completeness consistency between our reachability based zones and simulation results, as can be seen by the low number of false negative verification results.}
\label{table:completeness-experiment}
\end{table}

\subsection{Soundness Comparison}

While completeness is not compromised, we observe that maneuver-awareness increases soundness of our zones. This can be seen in Table~\ref{table:volume-comparison}, where we perform a comparison of state space volume between our maneuver zones and a baseline derivation in which the maneuver constraint is removed. This baseline zone effectively represents the union of all dynamically feasible maneuvers over the derivation time horizon. Our results show that we are able to achieve a 3-4x state space volume reduction compared to our baseline approach by considering maneuver constraints.

\begin{table}[h]
\centering
\begin{tabular}{ | c | c | }
    \hline 
    Zone Name & Volume Compared \\ 
       & to Baseline \\ 
    \hline 
    \hline
    Baseline & 100.0\% \\ 
    \hline
    Lane Change & 30.3\% \\  
    \hline
    $90^{\circ}$ Turn & 24.0\% \\  
    \hline 
\end{tabular}
\caption{A comparison of different obstacle perception zone volumes before and after maneuver-awareness is considered within the derivation. We observe a significant reduction in state space volume when applying maneuver-based decomposition to safety zones. In a V\&V setting, this implies a commensurate optimization in the performance requirement targets of an autonomous vehicle.}
\label{table:volume-comparison}
\end{table}

\subsection{Discussion}
Recall that the prototypical use case of safety zones is evaluation of an obstacle perception system. Depending on how a given system's safety-critical failures are distributed over state space, a reduction in zone volume implies a commensurate reduction in safety-critical failures which warrant investigation during V\&V \cite{Topan2022Interaction}. The significantly smaller, equally complete safety zones that we demonstrate in this work make stringent autonomous vehicle perception performance targets easier to achieve by providing a more precise definition of obstacle safety-criticality. 

We do note, however, that this improvement comes at the cost of necessitating a definition of each maneuver in terms of dynamic system constraints. Formulating a comprehensive V\&V strategy with this consideration will require all supported maneuvers of a system under test to be appropriately captured within the reachability formulation. 

A notable challenge of defining maneuvers using HJ-Reachability is expressing them using a small enough number of state variables to make computation tractable. Due to the curse of dimensionality, we found it difficult to model maneuvers requiring more than 7 state variables. Thus, modeling useful maneuvers requires tradeoffs between how prescriptive a maneuver can be and how feasible it is to express. We note, however, that the fallback of ignoring the intended maneuver and simply using a safety zone as in \cite{Topan2022Interaction} always exists; analysis of this tradeoff serves only to improve soundness. Looking ahead, certain research directions including grid-free solvers \cite{bansal2020deepreach} offer potential remedies to the described dimensionality issue.

\section{Conclusion}

In this work, we propose a maneuver-based decomposition of perception safety zones that leverages a novel temporal convolution operation with the capability to account for collision at any intermediate time along the way to maneuver completion. We demonstrate a significant reduction in zone volume while maintaining completeness, thus optimizing obstacle perception performance requirements by filtering out regions of state space not relevant to an AV's route.

An exciting direct application of this work lies in constructing a holistic risk assessment of an autonomous vehicle obstacle perception system, with a precise zone mapping for each maneuver context yielding adaptive, and overall less stringent, performance targets. Beyond reducing over-conservatism (i.e., improving soundness while maintaining completeness) in validating existing perception systems, these zones can also be used in developing such systems. For example, zone geometry and associated coverage analysis may be used to inform sensor placement, range requirements, or even to toggle sensor activation/field of view according to perception requirements in certain directions. For standard deep-learning-enabled perception systems, safety zones can also be used during training to weight the loss function to prioritize high recall within the zone. While we focus on autonomous vehicle maneuvers within this work, the approach we propose may be applied more broadly (i) to other perception-equipped cyber-physical systems operating in interactive environments such as autonomous drones or robotic arms, and (ii) to address more general temporal dependencies in reachability analysis, e.g., considering an accumulated comfort metric in addition to safety.

\newpage
\bibliographystyle{IEEEtran}
\bibliography{references}
\end{document}